\documentclass[letterpaper, 10 pt, conference]{ieeeconf}
\usepackage{times}
\usepackage[normalem]{ulem}
\usepackage{soul}
\usepackage{xcolor}
\usepackage{etoolbox}
\usepackage[pdftex]{graphicx}
\usepackage{amsmath,amssymb,amsopn,amstext,amsfonts}
\usepackage{cancel}
\usepackage[space]{cite}
\usepackage{balance}
\usepackage{color}
\usepackage{mathtools}
\usepackage{algpseudocode}
\usepackage{algorithm} 
\usepackage{bm}

\usepackage{diagbox}
\usepackage{float}
\usepackage{epstopdf}
\usepackage{pifont}
\usepackage{fixltx2e}
\usepackage{amsmath}
\usepackage{multirow}
\usepackage{url}
\usepackage{verbatim}
\usepackage[linkcolor=black,citecolor=black,urlcolor=black,colorlinks=true]{hyperref}
\usepackage{booktabs}
\usepackage{graphicx}
\usepackage{subcaption}
\usepackage{footmisc} 

\usepackage[normalem]{ulem}
\usepackage{xcolor}
\usepackage{etoolbox}

\usepackage{amsthm}

\algnewcommand\algorithmicforeach{\textbf{for each}}
\algdef{S}[FOR]{ForEach}[1]{\algorithmicforeach\ #1\ \algorithmicdo}
\newcommand{\tp}{^{\mathsf{T}}}

\newcommand{\delete}[1]{{\bgroup\markoverwith{\textcolor{red}{\rule[0.5ex]{2pt}{0.4pt}}}\ULon{#1}}}
\newcommand{\deletefig}[1]{{\bgroup\markoverwith{\textcolor{red}{\rule[2.5ex]{2pt}{2.0pt}}}\ULon{#1}}}

\graphicspath{{figures/}}
\DeclareGraphicsExtensions{.png,.jpg,.eps,.pdf}
\IEEEoverridecommandlockouts
\begin{document}
\title{TGK-Planner: An Efficient Topology Guided Kinodynamic Planner\\ for Autonomous Quadrotors}
\author{Hongkai Ye, Xin Zhou, Zhepei Wang, Chao Xu, Jian Chu and Fei Gao\thanks{All authors are with the State Key Laboratory of Industrial Control Technology, Institute of Cyber-Systems and Control, Zhejiang University, Hangzhou, 310027, China. {\tt\small Email:\{hkye, iszhouxin, wangzhepei, cxu\}@zju.edu.cn, chuj@iipc.zju.edu.cn and fgaoaa@zju.edu.cn}}}

\maketitle
\thispagestyle{empty}
\pagestyle{empty}

\begin{abstract}
In this paper, we propose a lightweight yet effective Topology Guided Kinodynamic planner (TGK-Planner) for quadrotor aggressive flights with limited onboard computing resources.
The proposed system follows the traditional hierarchical planning workflow, with novel designs to improve the robustness and efficiency in both the pathfinding and trajectory optimization sub-modules.
Firstly, we propose the \textit{topology guided graph}, which roughly captures the topological structure of the environment and guides the state sampling of a sampling-based kinodynamic planner.
In this way, we significantly improve the efficiency of finding a safe and dynamically feasible trajectory.
Then, we refine the smoothness and continuity of the trajectory in an optimization framework, which incorporates the homotopy constraint to guarantee the safety of the trajectory.
The optimization program is formulated as a sequence of quadratic programmings (QPs) and can be iteratively solved in a few milliseconds.
Finally, the proposed system is integrated into a fully autonomous quadrotor and validated in various simulated and real-world scenarios.
Benchmark comparisons show that our method outperforms state-of-the-art methods with regard to efficiency and trajectory quality.
Moreover, we will release our code as an open-source package\footnote{\label{footref:code}Code will be released after the acceptance of this paper at \url{https://github.com/ZJU-FAST-Lab/TGK-Planner}.}.
\end{abstract}


\IEEEpeerreviewmaketitle

\section{Introduction}
\label{sec:introduction}
In recent years, although many works have been proposed toward online aerial planning, it is still challenging for quickly generating high-speed kinodynamic trajectories in a resource-limited quadrotor.
Due to the complexity of the environment and system dynamics, generating an optimal and executable trajectory usually takes the price of high computational overhead.
Moreover, for a quadrotor flying at high speed, re-planning has to be finished in a short time to react to unpredictable obstacles.
For a cheap platform, especially the commercial quadrotor with a limited computing budget, the above two requirements are hard to be satisfied at the same time, making the high aggressiveness hard to achieve on the premise of safety guarantee.
Some works~\cite{lopez2017aggressive, ryll2019efficient} integrate perception with planning for high-speed flight by using motion primitive libraries. However, the restricted primitive set guarantees no optimality, and the discretization makes long-term trajectories inconsistent.

\begin{figure}[t]
\centering
\begin{subfigure}{0.77\linewidth}
	\includegraphics[width=1\linewidth]{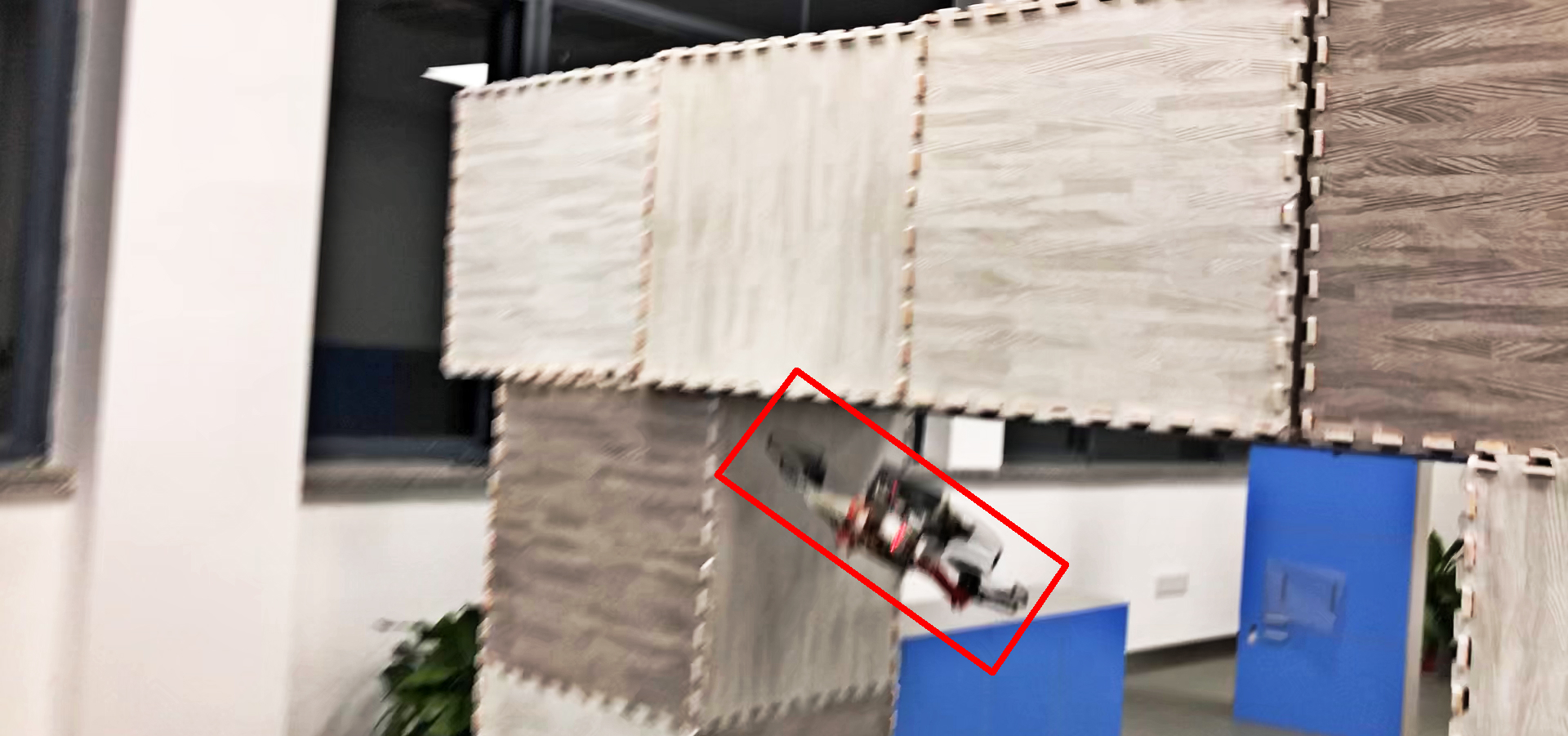}
	\caption{Dodge obstacles.}	
\end{subfigure}
\begin{subfigure}{0.77\linewidth}
	\includegraphics[width=1\linewidth]{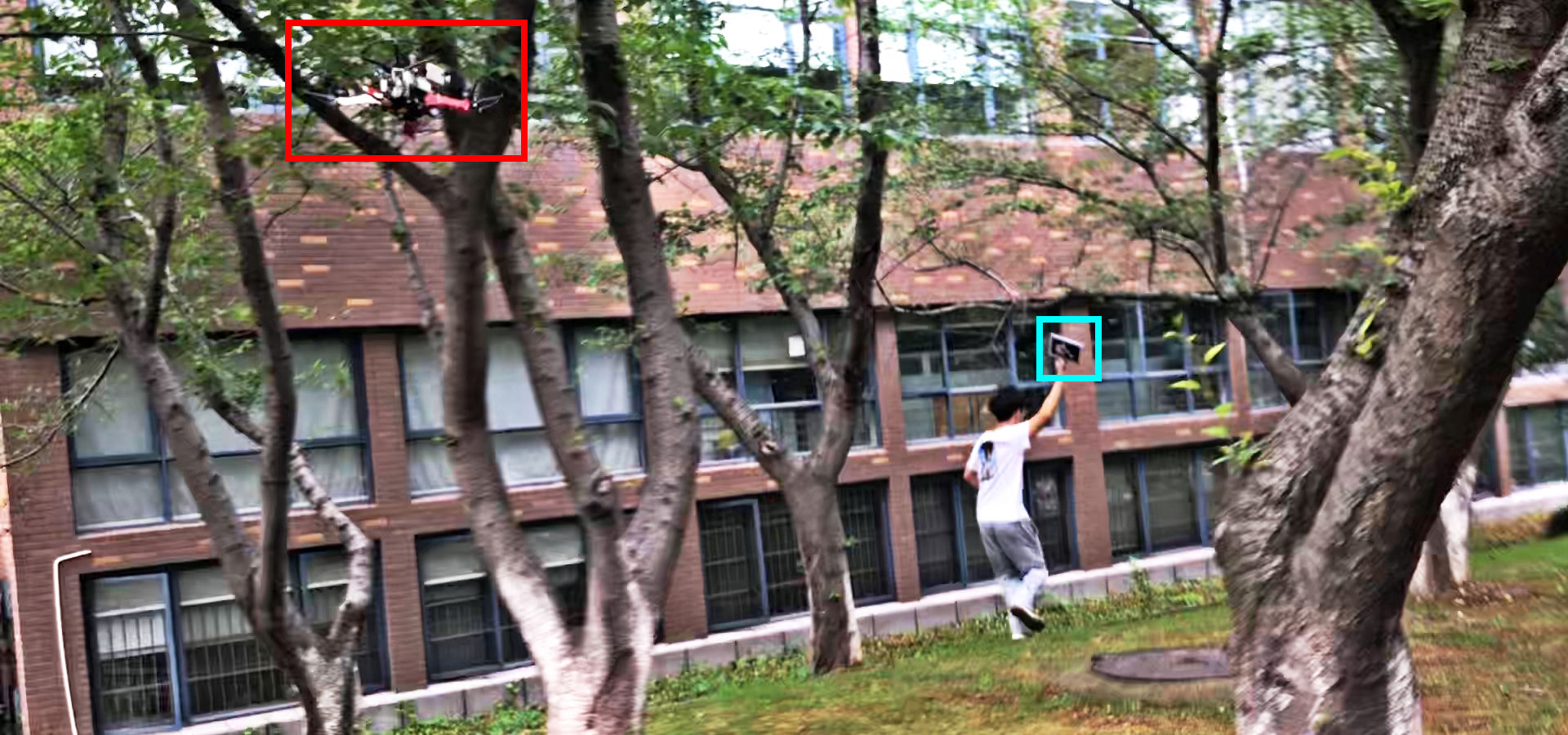}
	\caption{Chase a fast-moving target.}
	\label{fig:chase_qr}
\end{subfigure}
\captionsetup{font={small}}
\caption{Fast autonomous flight in unknown in(out)-door environments. Video is at \url{https://youtu.be/nNS0p8h5zAk}.}
\vspace{-0.5cm}
\end{figure}

In this paper, we investigate the above research gap and propose a systematic approach to bridge it.
Our method follows the traditional hierarchical planning workflow, which consists of a kinodynamic planner that finds a trajectory according to a coarse system dynamics, and an optimizer that improves the smoothness and continuity of the trajectory.
For kinodynamic planning in high-dimensional state spaces, sampling-based planners have great potential in efficiency by designing smart sampling strategies.
Imagine this situation: a quadrotor flies along a corridor at high speed, states sampled towards the walls are most probably not useful, while a state with a velocity along the corridor certainly benefits the planning.
Besides, many sampling-based methods have the \textit{anytime} nature, which especially suits fast flight by improving the optimality of the plan while executing it~\cite{karaman2011anytime}.

Therefore, we adopt a sampling-based front-end and efficiently sample states with environmental awareness.
Our front-end builds a topological structure capturing the free space's connectivity and then generates a high-quality feasible trajectory.
Based on this trajectory, we design a lightweight optimization-based back-end to further improve its key attributes, smoothness and continuity, with a guarantee on its safety and dynamical feasibility.
The proposed back-end fully exploits assets of our front-end, which are, reasonable homotopy residence and reasonable time allocation, by incorporating them into the objective.
Furthermore, efficiency and optimality are retained by formulating the optimization as a sequence of QPs with closed-form solutions.

This paper highlights its efficiency in both the front-end and back-end, guarantees the asymptotic optimality, and retains the robustness and quality of the generated trajectory.
We summarize the contributions as follow:

1) A sampling-based kinodynamic planning front-end, which significantly improves the efficiency of kinodynamic RRT*~\cite{webb2013kinodynamic} algorithm by environment guided state sampling, and the cost converges in a few milliseconds.

2) A lightweight yet effective trajectory refinement back-end, which exploits the front-end assets to improve the smoothness and continuity of the trajectory by a sequence of least-square optimization.

3) Integrating the proposed methods which suits both global and local trajectory generation into a fully autonomous quadrotor system, presenting extensive benchmark and experimental validations, and releasing source code for the reference of the community. 

\section{Related Work}
\label{sec:related_work}
\subsection{Kinodynamic Planning}
Kinodynamic planning can be roughly divided into search-based and sampling-based.
Search-based methods discretize the control space and use motion primitives to search for a solution with piece-wise constant controls.
Recent typical works~\cite{liu2017ral,boyu2019ral} develop efficient heuristics by solving an unconstrained linear-quadratic energy-time minimization problem.
However, in those methods, the resolution must be carefully chosen to make a trade-off between solution existence and search-space complexity. Besides, they leave apparent discontinuities in control inputs.
For sampling-based methods, RRT-based algorithms are naturally extendable to kinodynamic systems by sampling in the state space.
However, tree expansion can be extremely inefficient for complicated dynamics in high-dimensional state space.
This is mainly caused by inefficient boundary value problem (BVP) solving and invalid state sampling.
Webb et.al~\cite{webb2013kinodynamic} derives the closed-form solutions to solve the BVP for linearized systems with a nilpotent dynamics matrix, which saves the computational overhead.
Nevertheless, too much computation is wasted on connecting invalid samples, making it impossible for real-time usage on embedded platforms.
To increase the probability of obtaining valid samples, it is necessary to design a strategy to bias/guide the sampling process.
Some works\cite{Cover2013sparse, blochliger2017topomap, oleynikova2018sparse} build sparse skeleton graphs of the environment and generate samples alongside edges of the graph.
These works only consider cases in the $\mathbb{R}^3$ space, and extracting a complete topological graph is rather time-consuming as the scale and complexity of the environment grow.
In this paper, we build our front-end upon~\cite{webb2013kinodynamic} and propose a simple yet effective topology extraction method to guide the sampling.

\subsection{Trajectory Optimization}
Trajectory optimization is essential in improving the path found by the front-end to meet the full system dynamics.
The minimum-snap formulation~\cite{MelKum1105} is widely adopted due to its simplicity and efficiency.
In~\cite{RicBryRoy1312}, the authors further convert it to an unconstrained quadratic programming (QP) problem and solve it in closed-form.
The safety and dynamic feasibility of the trajectory is ensured by iteratively adding intermediate waypoints to the path and solving the QP.
Some works~\cite{liu2017ral}~\cite{fei2018jfr,gao2019teach,tordesillas2019faster} extract obstacle-free corridors represented by a sequence of convex enclosed shapes, and then generate safe trajectories within the free space by convex optimization.
Although these works enjoy the convexity in their formulations, too many hard-constraints impose intensive computational overhead, thus preventing them from being used in cheap platforms.
Besides, no dynamics is considered in their front-end, making the optimization process always over-conservative.
Gradient-based methods~\cite{ratliff2009chomp,oleynikova2016continuous,fei2017iros} formulate trajectory optimization as non-linear optimization problems with penalties on collision, control, and constraint violation.
For ensuring safety, a costly Euclidean signed distance field (ESDF) has to be established, and the integration of cost terms is usually expensive.
Some recent works~\cite{Usenko2017ewok,boyu2019ral} mitigate these issues by parameterizing the trajectory as B-splines, and aggregate costs only on discrete control points.
However, due to the underlying nonlinearity of the optimization program, it can not guarantee a good final solution and is sensitive to the initial guess.
In this paper, we formulate our optimization problem as a sequence of QPs and utilize the topological information from the front-end, to design a fast and robust optimization pipeline.

\section{Kinodynamic Trajectory Planning}
\label{sec:front_end}
We briefly review the Kinodynamic RRT*~\cite{webb2013kinodynamic} algorithm, and then present our environment guided sampling strategy which significantly facilitates the efficiency.

\begin{figure}[t]
\centering
\begin{subfigure}{0.49\linewidth}
	\includegraphics[width=1\linewidth]{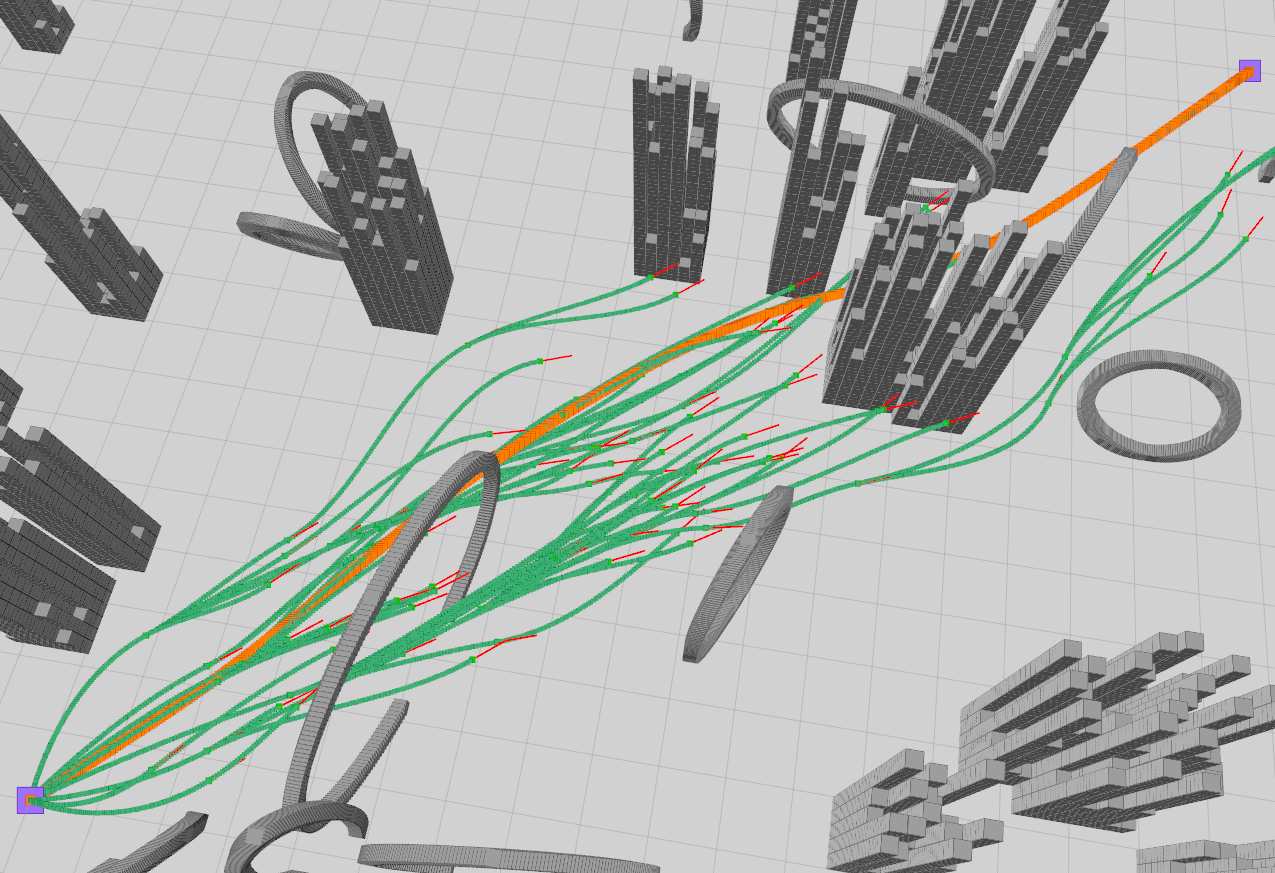}
	\caption{Proposed sampling strategy}
\end{subfigure}
\begin{subfigure}{0.49\linewidth}
	\includegraphics[width=1\linewidth]{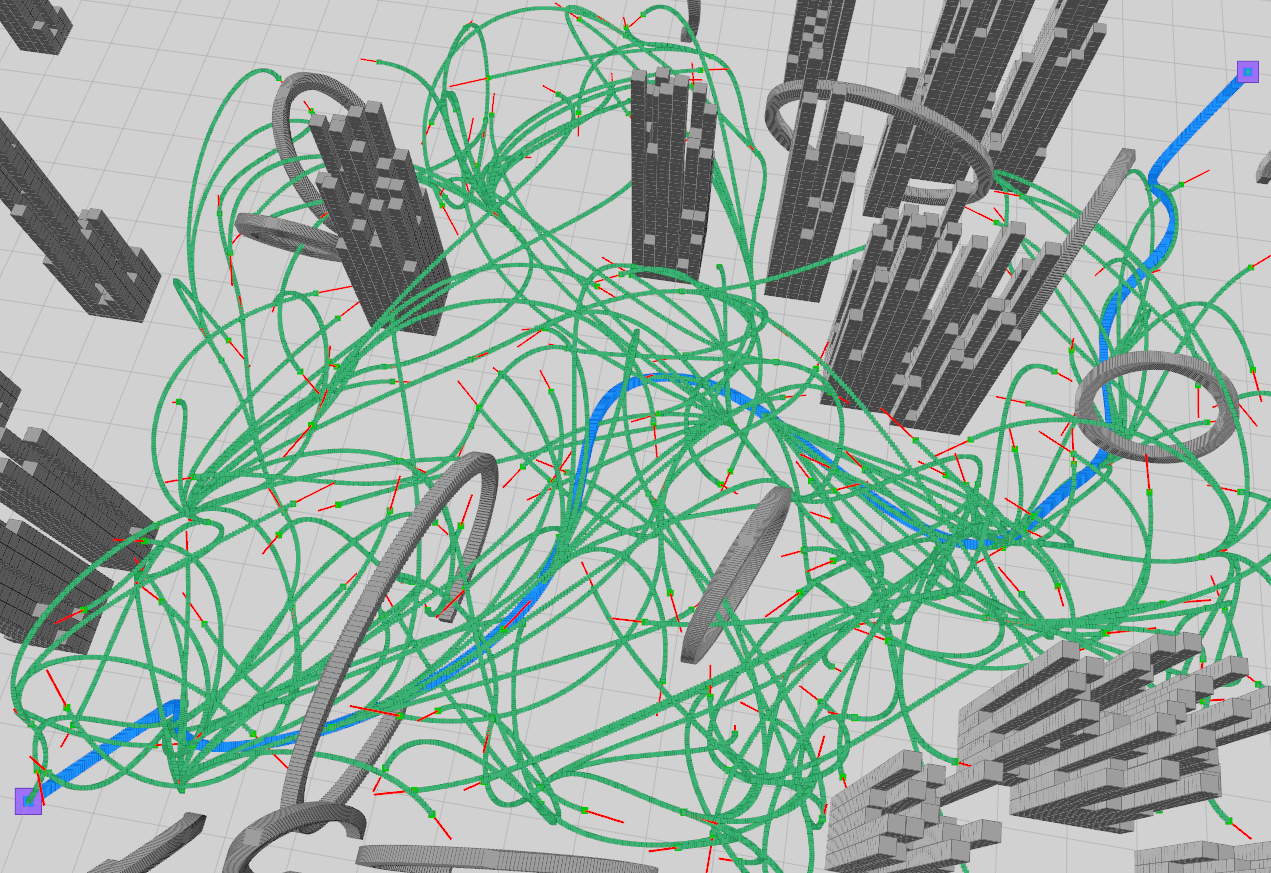}
	\caption{Uniformly random sampling}	
\end{subfigure}
\captionsetup{font={small}}
\caption{a) The generated trees (green) and the first feasible trajectory found by our guided sampling strategy (orange) in 3ms; and b) by uniformly random sampling (blue) in 100ms.}
\label{pic:guided_vs_random}
\vspace{-0.5cm}
\end{figure}

\subsection{Kinodynamic RRT* Framework}
The main workflow of the Kinodynamic RRT*~\cite{webb2013kinodynamic} is described in Alg.~\ref{alg:kino_rrt_star}, where a tree $\mathcal{T}$ grows from the initial state $\mathbf{x}_{init}$ towards the goal state $\mathbf{x}_{goal}$.
Every time a valid state $\mathbf{x}_{random}$ is sampled, a subset of $\mathcal{T}$, $\mathcal{X}_{backward}$ whose elements can connect to $\mathbf{x}_{random}$ are found though \textbf{BackwardNear()}.
If $\mathcal{X}_{backward}$ is not empty, then a node $\mathbf{x}_{min}$ with the minimum transition cost is chosen as the parent node of $\mathbf{x}_{random}$ through \textbf{ChooseParent()}, and $\mathbf{x}_{random}$ is added to the tree $\mathcal{T}$.
Moreover, \textbf{ForwardNear()} searches in $\mathcal{T}$ for a node set $\mathcal{X}_{forward}$ whose elements that $\mathbf{x}_{random}$ can connect to, and then \textbf{Rewire()} checks for every state in $\mathcal{X}_{forward}$ whether it can be reached by a lower cost route though $\mathbf{x}_{random}$.
The loop terminates when either the maximum sampling number or the running time exceeds.
Finally, the trajectory is obtained by tracing back from $\mathbf{x}_{goal}$ through its parent recursively, if $\mathbf{x}_{goal}$ is connected with any state node in the tree.
A visualization is provided in Fig.~\ref{pic:guided_vs_random}.

\begin{algorithm}[t]
\caption{Kinodynamic RRT*}
\label{alg:kino_rrt_star}
\begin{algorithmic}[1]
\State \textbf{Notation}: Environment $\mathcal{E}$, Tree $\mathcal{T}$, State $\mathbf{x}$
\State Initialize: $\mathcal{T} \leftarrow \emptyset \cup \{\mathbf{x}_{init}\}$
\For{$i = 1$ to $n$}
    \State $\mathbf{x}_{random} \leftarrow$ \textbf{Sample}($\mathcal{E}$)
    \State $\mathcal{X}_{backward} \leftarrow$ \textbf{BackwardNear}($\mathcal{T}$, $\mathbf{x}_{random}$)
    \If{$\mathcal{X}_{backward} \not= \emptyset$}
        \State $\mathbf{x}_{min} \leftarrow$ \textbf{ChooseParent}($\mathcal{X}_{backward}$, $\mathbf{x}_{random}$)
        \State $\mathcal{T} \leftarrow \mathcal{T} \cup \{\mathbf{x}_{random}\}$
        \State $\mathcal{X}_{forward} \leftarrow$ \textbf{ForwardNear}($\mathcal{T}$, $\mathbf{x}_{random}$)
        \If{$\mathcal{X}_{forward} \not= \emptyset$}
            \State \textbf{Rewire}($\mathcal{T}$, $\mathcal{X}_{forward}$)
        \EndIf
    \EndIf
\EndFor
\State \Return $\mathcal{T}$
\end{algorithmic}
\end{algorithm}

\subsection{Optimal States Transition}
\label{subsec:optimal_connect}
The cornerstone in the above Alg.~\ref{alg:kino_rrt_star} is the optimal connection of two states.
In~\cite{webb2013kinodynamic}, a general form of the connection for systems with a nilpotent dynamics matrix and a mixed time/energy cost criterion is derived.
Specifically, we fit it to our model and derive optimal solutions using standard optimal control techniques.
In this paper, the transition cost from state $\mathbf{x}_0$ to state $\mathbf{x}_1$ is defined as:
\begin{equation}
\label{equ:cost_func}
c(\mathbf{x}_0, \mathbf{x}_1)=\int_{0}^{\tau}(\rho+\frac{1}{2}\mathbf{u}(t)\tp\mathbf{u}(t))dt,
\end{equation}
where $\tau$ is the time duration and $\rho$ is the weight.
Minimizing the cost is equivalent to solve a fixed-endpoint, free-time optimal control problem \cite{liberzon2012calculus}:
\begin{equation}
\label{equ:problem_form}
\begin{split}
\min \mathcal{J}&(\mathbf{x}(t))=\int_{0}^{\tau}\mathcal{L}(t, \mathbf{x}(t), \mathbf{\dot{x}}(t), \mathbf{u}(t))dt \\[1ex]
s.t. \quad & {f}(t, \mathbf{x}, \mathbf{u}) - \mathbf{\dot{x}}(t)=\mathbf{0}, \\
&\mathbf{x}(0)=\mathbf{x}_0, \
\mathbf{x}(\tau)=\mathbf{x}_1, \\
&\mathbf{x}(t) \in \mathcal{X}^{free}, \
\mathbf{u}(t) \in \mathcal{U}^{free},
\end{split}
\end{equation}
where Lagrangian $\mathcal{L}$ is the cost functional defined in Eq.~\ref{equ:cost_func}, and ${f}(t, \mathbf{x}, \mathbf{u})$ is the differential constraint of the system:

\begin{equation}
\label{equ:dynamics}
\mathbf{\dot{x}}(t)=\mathbf{A}\mathbf{x}(t)+\mathbf{B}\mathbf{u}(t),
\end{equation}
\begin{equation}
\begin{split}
&\mathbf{x}(t)=
\begin{bmatrix}
\mathbf{p}(t) \\
\mathbf{\dot{p}}(t)
\end{bmatrix}, \quad
\mathbf{A}=
\begin{bmatrix}
\mathbf{0} & \mathbf{I} \\
\mathbf{0} & \mathbf{0}
\end{bmatrix}, \quad
\mathbf{B}=
\begin{bmatrix}
\mathbf{0} \\
\mathbf{I}
\end{bmatrix}, \\[1ex]
&\mathbf{u}(t)=\mathbf{\ddot{p}}(t), \quad
\mathbf{p}(t) =
\begin{bmatrix}
p_x(t),\ p_y(t),\ p_z(t)
\end{bmatrix}\tp,
\end{split}
\end{equation}
which is modeled as a linear system according to the quadrotor's differential flatness property\cite{MelKum1105}.

According to the calculus of variation, the Hamiltonian is written as
$\mathcal{H}(t, \mathbf{x}, \mathbf{u}, \mathbf{\lambda})=\mathcal{L}+\mathbf{\lambda}\tp \mathbf{f}$,
where $\mathbf{\lambda}(t)$ is the costate vector.
In our case, the optimal arriving time $\tau^*$ satisfies $\mathcal{H}(\tau^*, \mathbf{x}, \mathbf{u}, \mathbf{\lambda})=0$, which is an equation of $4^{th}$ order polynomial whose coefficients are fully determined by boundary conditions.
After solving this equation, $\tau^*$ is obtained and the problem becomes a fixed-endpoint, fixed-time problem.

Let $\mathbf{u}^*$ and $\mathbf{x}^*$ be the optimal control and state trajectory separately, we now apply Pontryagin Maximum Principle \cite{liberzon2012calculus} to characterize $\mathbf{u}^*$. The state $\mathbf{x}^*$ and costate $\mathbf{\lambda}^*$ must satisfy the following canonical equations:
\begin{equation}
\begin{cases}
\dot{\mathbf{\lambda}}^*=-{\partial \mathcal{H}(t, \mathbf{x}^*, \mathbf{u}^*, \mathbf{\lambda}^*)} / {\partial \mathbf{x}}, & \\
\dot{\mathbf{x}}^*={\partial \mathcal{H}(t, \mathbf{x}^*, \mathbf{u}^*, \mathbf{\lambda}^*)} / {\partial \mathbf{\lambda}}. & \\
\end{cases}
\end{equation}

If assuming the control and state unbounded, the maximizer of $\mathcal{H}$ satisfies ${\partial \mathcal{H}} / {\partial \mathbf{u}^*}=\mathbf{0}$.
Solving this equation along with the boundary and transversality conditions, we obtain the optimal solution pair $\{\mathbf{u}^*(t)$, $\mathbf{p}^*(t)\}$ which is:
\begin{equation}
\label{equ:p_u}
\begin{split}
&p_{k}^*(t)=\frac{1}{6}c_{k,3}t^3+\frac{1}{2}c_{k,2}t^2+c_{k,1}t+c_{k,0}, \\
&u_{k}^*(t)=c_{k,3}t+c_{k,2}, \ k \in \{x, y, z\}.\\[1ex]
\end{split}.
\end{equation}

The corresponding optimal cost can be derived from Eq.~\ref{equ:cost_func}.
We then check the feasibility of the unconstrained optimal solution pair $\{\mathbf{u}^*(t)$, $\mathbf{p}^*(t)\}$, and consider the connection failed if it violates any constraints.
Note, although this simplification of the maximizer of $\mathcal{H}$ sacrifices some feasible samples, it greatly accelerates the BVP solving and facilities the overall efficiency.

\begin{figure}[t]
\centering
\begin{subfigure}{0.85\linewidth}
	\includegraphics[width=1\linewidth]{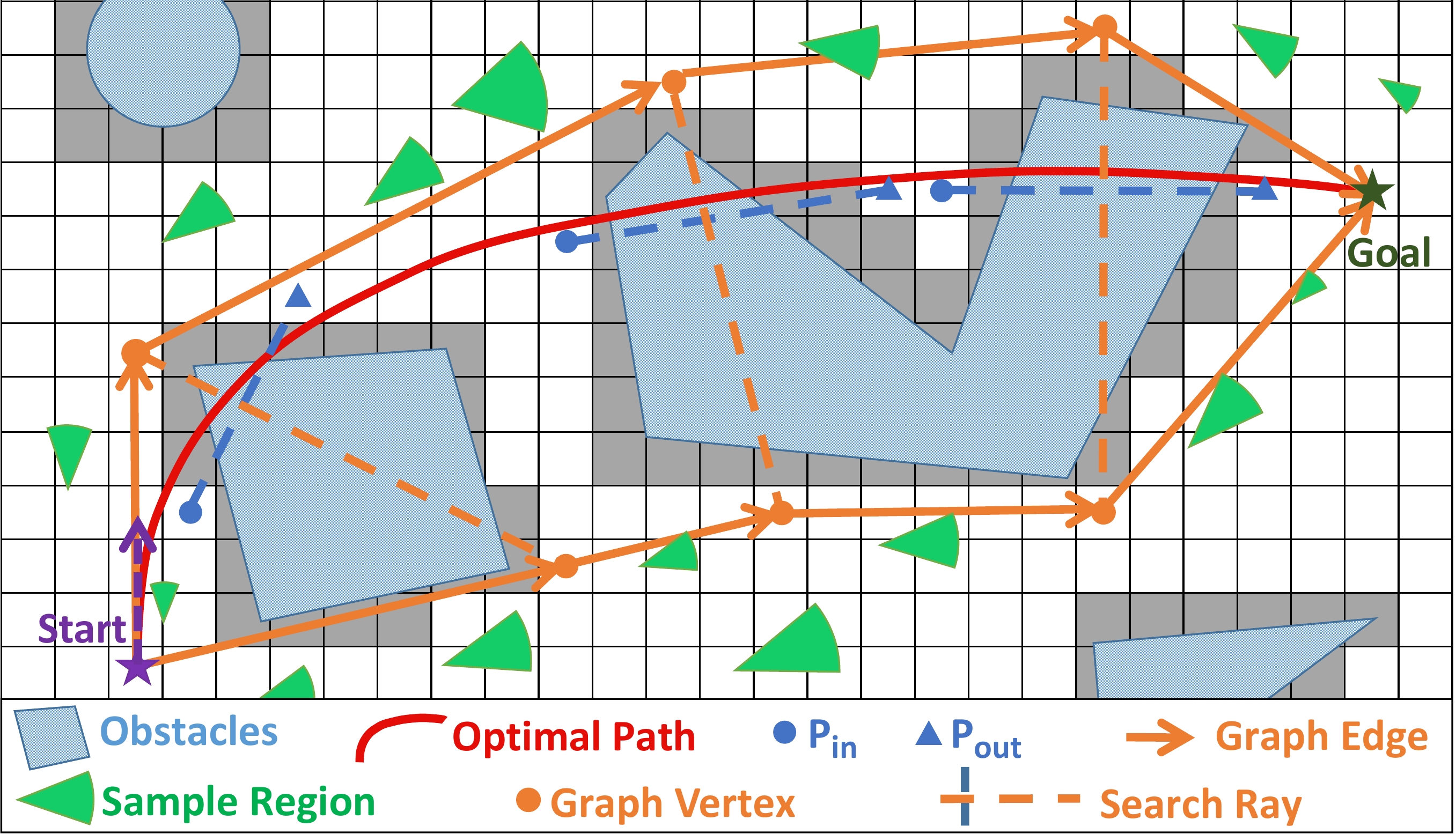}	
	\caption{}
\end{subfigure}
\begin{subfigure}{0.48\linewidth}
	\includegraphics[width=1\linewidth]{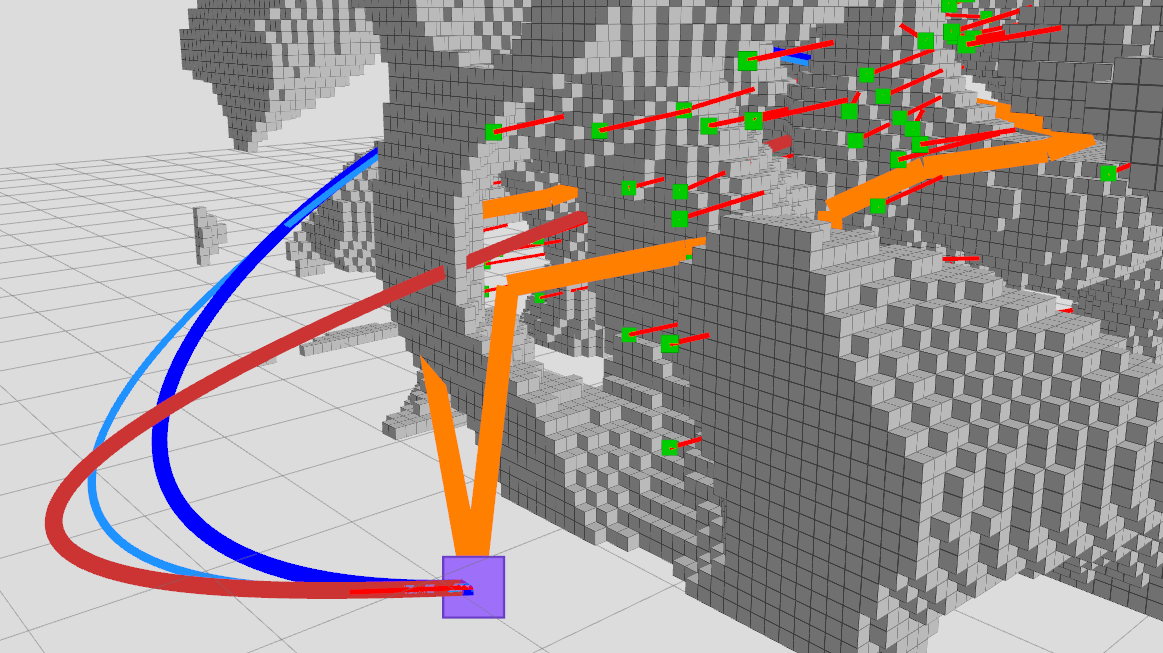}	
	\caption{}
\end{subfigure}
\begin{subfigure}{0.45\linewidth}
	\includegraphics[width=1\linewidth]{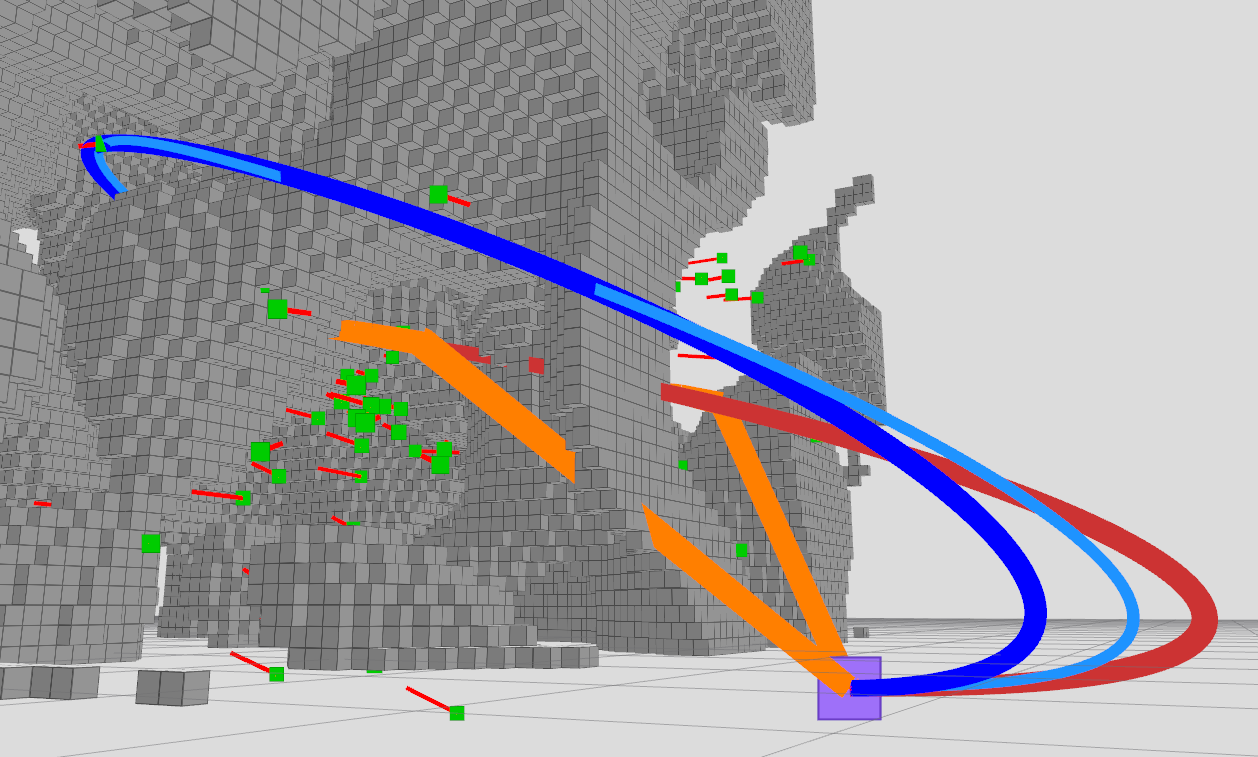}	
	\caption{}
\end{subfigure}
\captionsetup{font={small}}
\caption{(a) Illustration of constructing a topological graph. Orange lines form the graph and green fans represent the probabilistic-based sampling region of states. (b)(c) Graph (Orange) and state samples (Green dots represent state position and red lines represent state velocity) in 3D environments. The initial velocity is nonzero.}
\label{pic:topo}
\vspace{-0.3cm}
\end{figure}

\subsection{Approximate Topological Graph Guided Sampling}
\label{subsec:sample}

As intuitively stated in Sect.~\ref{sec:introduction}, a uniformly random sampling of the free states is inefficient.
We here use a method to quickly construct a \textit{topology guided graph}, which approximately captures the topological structure of the environment, as shown in Fig~\ref{pic:topo}.
The environment is represented as an occupancy grid map.
To construct the \textit{graph}, an optimal path directly connecting $\mathbf{x}_{init}$ and $\mathbf{x}_{goal}$ is firstly planned without considering any obstacles (Fig.~\ref{pic:topo}, red curve).
Along the path, we record positions where the path goes in and out of obstacles, denoting as $\mathbf{p}_{in}^{i}$ and $\mathbf{p}_{out}^{i}$.
Connecting each pair of them forms \textit{traversal lines} (Fig.~\ref{pic:topo}, dashed blue line).
Then, starting from the middle point of each \textit{traversal line}, we do ray tracing (Fig.~\ref{pic:topo}, dashed orange line) in the direction perpendicular to the \textit{traversal line} and level to the horizontal plane.
The tracing stops when an obstacle-free grid is found on both sides, and the stopping grids are taken as vertices of the \textit{graph}, which are of the same height as the middle point of the corresponding \textit{traversal line}.
Positions of $\mathbf{x}_{init}$ and $\mathbf{x}_{goal}$ are also graph vertices.
Finally, the \textit{graph} is constructed by connecting the vertices from start to goal.
Unlike ~\cite{simon2000visibility} and other methods that desire complete topological graphs in obstacle-free areas, our graph captures the partial topological structure of the environment in a much cheaper way.
As a sacrifice, the graph edges (Fig.~\ref{pic:topo}, solid orange line) are not guaranteed collision-free.
However, this is acceptable since we sample state positions in free space in the vicinity of these edges with a normal distribution, as shown in Fig.~\ref{pic:topo}.
As for state velocity, its direction is sampled with a normal distribution to deviate from the direction of edges. Its magnitude is sampled according to speed limits.

\section{Fast Trajectory Refinement}
\label{sec:back_end}
As stated before, a trajectory obtained from the front-end (Sect.~\ref{sec:front_end}) is based on a coarse dynamic model and, therefore, has relatively low fidelity despite it meets all constraints.
In this section, we show how to efficiently improve the continuity and smoothness, by incorporating the homotopy structure of the front-end trajectory.

\subsection{Problem Formulation}
\label{subsec:back_end_formulation}
For each dimension, consider an $m$-segment, $n^{th}$-order polynomial trajectory $p_{m}(t)=c_0+c_1t+c_2t^2\cdots+c_nt^n,$
and let $\mathbf{c}_{m}=[c_0,c_1,c_2,\cdots,c_n]\tp$ be the coefficient of the $m^{th}$ segment, our goal is to find the optimal coefficient for each segment of the trajectory.


To optimize the trajectory, we investigate the proposed front-end, and build our back-end based on some special properties of its solution.
Firstly, the quadrotor dynamics is roughly captured in the front-end, making the initial path be in a reasonable homotopy class (geometric region).
As proved by~\cite{liu2017ral,dolgov2010path}, a much better trajectory can be obtained starting from this initial trajectory and search in its nearby solution space.
Secondly, the trajectory satisfies all the constraints imposed by the acceleration input model, including safety constraints and dynamical constraints.
It is $C^0$ and $C^1$ continuous but only segment-wise $C^2$ continuous, that is, 
the acceleration changes abruptly in conjoined points between every two consecutive segments, although it is continuous within each segment. We define the acceleration differences between segments as an acceleration gap and aim to minimize it in the following optimization progress since the gap leads to quadrotor attitude jitters, which harm the control a lot.

Based on the above observations, we let the objective $J$ make out of three terms, and the problem becomes:
\begin{equation}
\begin{split}
\min \ &J =\ \lambda_{s} J_{s}+\lambda_{h} J_{h}+\lambda_{c} J_{c}\\[1ex]
s.t. \quad &\mathbf{x}(t) \in \mathcal{X}^{free}, \
\mathbf{u}(t) \in \mathcal{U}^{free},
\end{split}
\end{equation}
where $J_{s}$ is the cost of overall smoothness, $J_h$ the term that penalizes the difference in homotopy class compared with the front-end trajectory, $J_c$ the term that penalizes acceleration discontinuity between segments, and $\lambda_{s}$, $\lambda_{h}$, $\lambda_{c}$ the weights.

\begin{figure}[t]
\centering
\begin{subfigure}{0.4\linewidth}
	\includegraphics[width=1\linewidth]{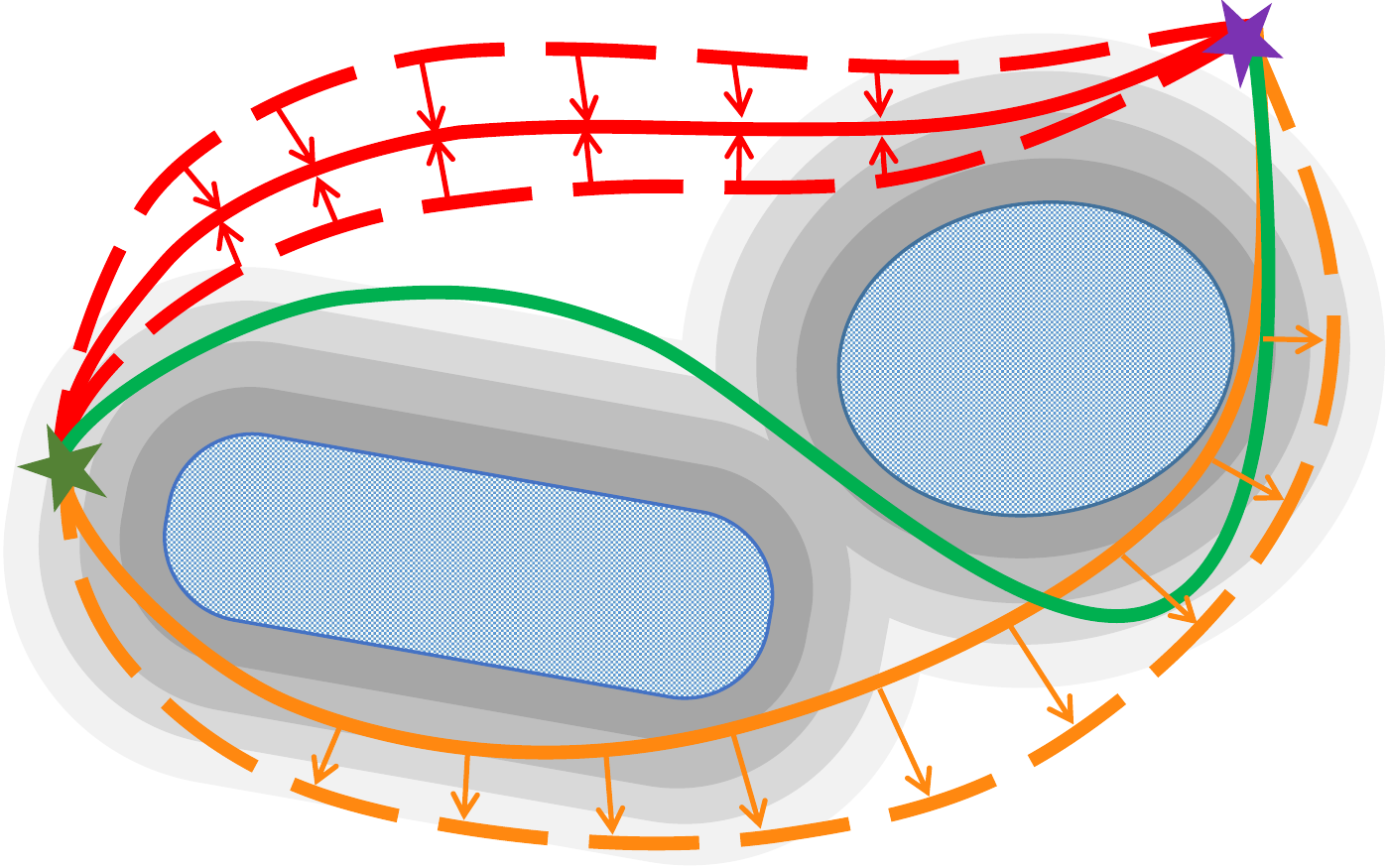}	
	\caption{}
	\label{subfig:attract}
\end{subfigure}
\begin{subfigure}{0.4\linewidth}
	\includegraphics[width=1\linewidth]{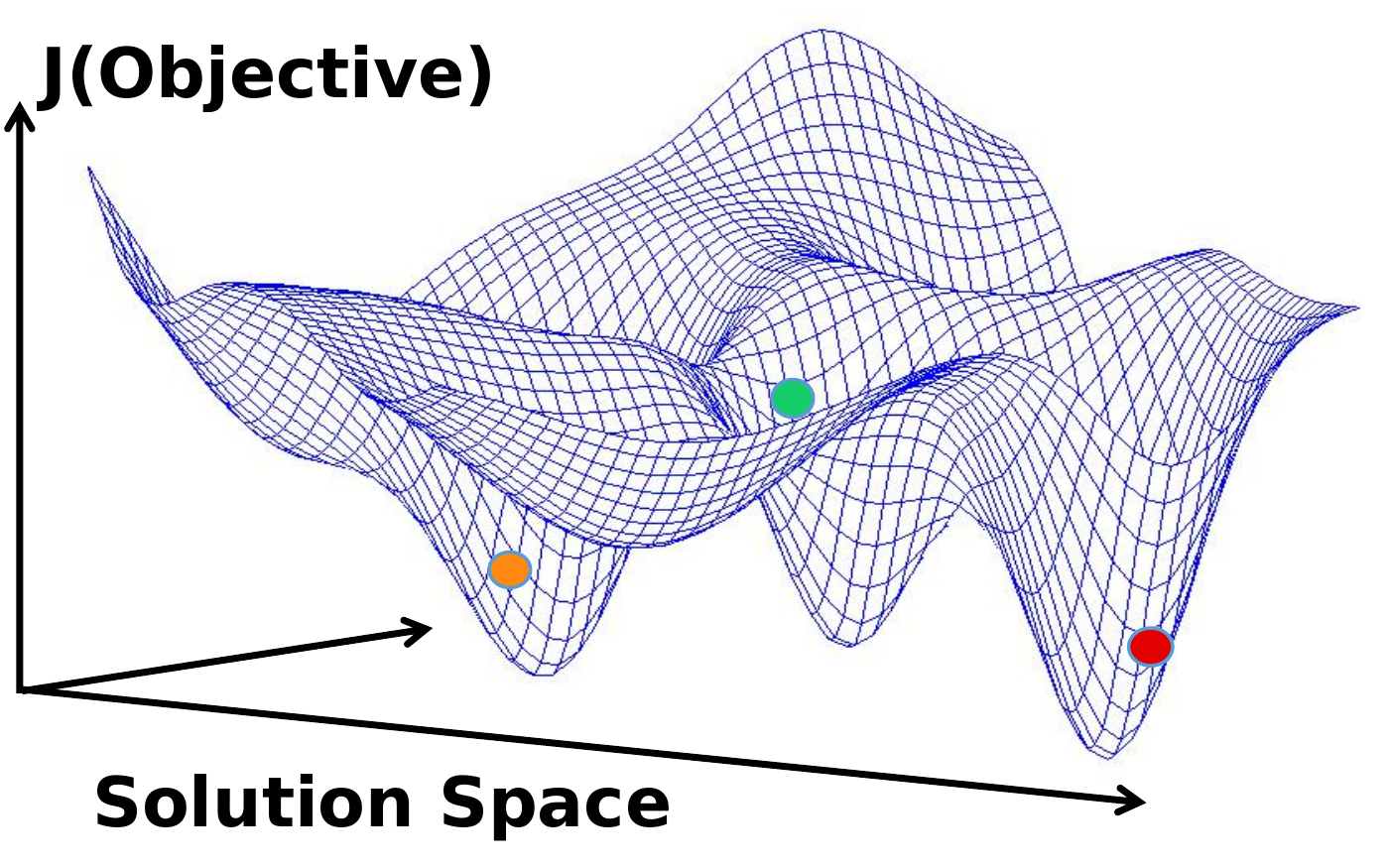}
	\caption{}
	\label{subfig:pits}
\end{subfigure}
\vspace{-0.2cm}
\captionsetup{font={small}}
\caption{Illustration of the homotopy constraint. (a) In our method, the initial path (solid red curve) attracts optimized paths (red dashed curves). In many others, an ESDF pushes path (orange curves) away from obstacles. (b) For a highly nonconvex optimization, solutions of different homotopy classes fall in different "pits" (nearby solution spaces of local minima) of the objective. }
\label{pic:topo_pits}
\vspace{-0.4cm}
\end{figure}

Here, the homotopy penalty $J_{h}$ is essential, since it makes the online optimization with the above highly nonlinear constraints solvable.
By adding this term, we turn the safety constraint from a collision rejecting one to a feasible solution attracting one, and avoid the expensive computation for an ESDF, as shown in Fig.~\ref{subfig:attract}.
Besides, it significantly narrows the alternative solutions to a nearby solution space of the initial feasible solution, as shown in Fig.~\ref{subfig:pits}.
Based on all these above, we design our optimization framework as a homotopy penalized, soft-constrained, iterative optimization problem.
Fortunately, since all cost terms are quadratic, each iteration of the optimization has a closed-form optimal solution that is efficient and numerically stable.

\subsection{Quadratic Objective Construction}
\subsubsection{Smoothness Cost}
$J_s$ is formulated as the integral of the squared derivative of the trajectory:
\begin{equation}
\begin{aligned}
J_s=&\sum_{k \in \{x, y, z\}} \int_0^T[p_{k}^{(j)}(t)]^2dt \\
=&\sum_{k \in \{x, y, z\}} \sum_{i=1}^{m} \mathbf{c}_{i,k}\tp \int_0^{t_i}\mathbf{t}^{(j)} (\mathbf{t}^{(j)})\tp dt \ \mathbf{c}_{i,k} \\
=&\sum_{k \in \{x, y, z\}} \mathbf{c}\tp \mathbf{Q}_s \mathbf{c},
\end{aligned}
\end{equation}
where $T=t_1+t_2+\cdots+t_m$ is the total duration of the trajectory and $t_i$ the duration for each segment.
$\mathbf{t}^{(j)}=\mathrm{d}^j[1,t,t^2,\cdots,t^n]\tp / \mathrm{d}t^j$ is the $j^{th}$-order derivative vector of $\mathbf{t}=[1,t,t^2,\cdots,t^n]\tp$, and $\mathbf{c}\tp=[\mathbf{c}_{1,k}\tp, \mathbf{c}_{2,k}\tp, \cdots, \mathbf{c}_{m,k}\tp]$ is the coefficient vector of $m$ segments.

\subsubsection{Homotopy Cost}
The homotopy cost $J_h$ is formulated as the integration over the squared difference between positions of the optimized trajectory and the original trajectory:
\begin{equation}
\begin{aligned}
J_h=&\sum_{k \in \{x, y, z\}} \int_0^T[p_k(t)-p_k^*(t)]^2dt \\
=&\sum_{k \in \{x, y, z\}} \sum_{i=1}^{m} (\mathbf{c}_{i,k}-\mathbf{c}_{i,k}^*)\tp \int_0^{t_i}\mathbf{t} \mathbf{t}\tp dt \ (\mathbf{c}_{i,k}-\mathbf{c}_{i,k}^*) \\
=&\sum_{k \in \{x, y, z\}} (\mathbf{c}-\mathbf{c}^*)\tp \mathbf{Q}_h (\mathbf{c}-\mathbf{c}^*),
\end{aligned}
\end{equation}
where $p^*(t)$ is the original trajectory with $\mathbf{c}^*$ its coefficient vector of $m$ segments.

By adding this term, the optimizer will force the optimized trajectory to be close to the original one, thus more likely to be residing in free spaces of the same homotopy class.

\subsubsection{Continuity Cost}
We penalize the acceleration gap for approaching near $C^2$ continuity. The continuity cost is defined as:
\begin{equation}
\begin{aligned}
J_c=&\sum_{k \in \{x, y, z\}} \sum_{i=1}^{m-1}[\ddot{p}_{i,k}(t_i)-\ddot{p}_{i+1,k}(0)]^2 \\
=&\sum_{k \in \{x, y, z\}} \sum_{i=1}^{m-1}[\mathbf{c}_{i,k}\tp \mathbf{t}^{(2)}\lvert_{t=t_i}-\mathbf{c}_{i+1,k}\tp \mathbf{t}^{(2)}\lvert_{t=0}]^2 \\
=&\sum_{k \in \{x, y, z\}} \mathbf{c}\tp \mathbf{Q}_c\mathbf{c},
\end{aligned}
\end{equation}
where $\ddot{p}_{i,k}(t_i)$ is the terminal acceleration of the $i^{th}$ segment and $\ddot{p}_{i+1,k}(0)$ is the beginning acceleration of the $(i+1)^{th}$ segment, both in the in $k$ dimension.

Here, we formulate the acceleration gap penalty as a soft constraint, since imposing a hard constraint of overall $C^2$ continuity may prevent finding a feasible solution, especially among extremely cluttered obstacles.
Considering safety as the top priority for planning, a minor acceleration gap is acceptable in exchange for higher possibilities to find trajectories with strict safety guarantees.

With the terms mentioned above, the overall objective function is written in a quadratic form:
\begin{equation}
\begin{aligned}
\min J =&\ \lambda_{s} J_{s}+\lambda_{h} J_{h}+\lambda_{c} J_{c} \\[1ex]
=&\sum_{k \in \{x, y, z\}} [\mathbf{c}\tp(\lambda_{s}\mathbf{Q}_s+\lambda_{h}\mathbf{Q}_h+\lambda_{c}\mathbf{Q}_c)\mathbf{c}- \\
& 2\lambda_{h}\mathbf{c}\tp \mathbf{Q}_h\mathbf{c}^*+\lambda_{h}(\mathbf{c}^*)\tp \mathbf{Q}_h\mathbf{c}^*] \\[1ex]
s.t. \quad \mathbf{A}&\mathbf{c}=\mathbf{d},
\end{aligned}
\end{equation}
where $\mathbf{c}$ is the decision variable, and $\mathbf{A}\mathbf{c}=\mathbf{d}$ is the boundary derivative constraints for each segments.
The cost is independent of each axis and can be solved separately.


\subsection{Closed-form Solution for Each Iteration}
As described in \cite{RicBryRoy1312}, a piecewise polynomial trajectory can be expressed in term of boundary derivatives instead of coefficients of each segment:
\begin{equation}\label{equ:cKd}
\mathbf{c}=\mathbf{K}
\begin{bmatrix}
\mathbf{d}_f \\
\mathbf{d}_p
\end{bmatrix}, \
\mathbf{K}=\mathbf{A}^{-1}\mathbf{C},
\end{equation}
where matrix $\mathbf{K}$ maps the coefficients vector $\mathbf{c}$ to the derivatives vector which is reordered as fixed derivatives $\mathbf{d}_f$ and free derivatives $\mathbf{d}_p$ (the decision variables).
Details about the construction of the mapping matrix are described in \cite{RicBryRoy1312}.

In this way, the objective can be rewritten in an unconstrained formulation in each dimension as:
\begin{equation}
\begin{aligned}
J =&
\begin{bmatrix}
\mathbf{d}_f \\
\mathbf{d}_p
\end{bmatrix}\tp \mathbf{K}\tp (\lambda_{s}\mathbf{Q}_s+\lambda_{h}\mathbf{Q}_h+\lambda_{c}\mathbf{Q}_c)\mathbf{K}
\begin{bmatrix}
\mathbf{d}_f \\
\mathbf{d}_p
\end{bmatrix} -\\[1ex]
&2\lambda_h
\begin{bmatrix}
\mathbf{d}_f \\
\mathbf{d}_p
\end{bmatrix}\tp \mathbf{K}\tp \mathbf{Q}_h \mathbf{c}^*+
\lambda_h (\mathbf{c}^*)\tp \mathbf{Q}_h \mathbf{c}^*.
\end{aligned}
\end{equation}

Denote $\mathbf{K}\tp (\lambda_{s}\mathbf{Q}_s+\lambda_{h}\mathbf{Q}_h+\lambda_{c}\mathbf{Q}_c)\mathbf{K}$ as matrix $\mathbf{R}$, $\mathbf{K}\tp \mathbf{Q}_h \mathbf{c}^*$ as matrix $\mathbf{Z}$. Omit constants in $J$ which do not affect the optimal solution, the Jacobian of $J$ with respect to $\mathbf{d}_p$ in one axis is:

\begin{equation}
\frac{\partial J}{\partial \mathbf{d}_p}=2\mathbf{R}_{pf}\mathbf{d}_f+2\mathbf{R}_{pp}\mathbf{d}_p-2\lambda_{h}\mathbf{Z}_p,
\end{equation}
where $\mathbf{R}_{xx}$ and $\mathbf{Z}_x$ are block matrices of $\mathbf{R}$ and $\mathbf{Z}$.
Let the Jacobian equal $\mathbf{0}$, and we get the closed-form solution of the decision variables:
\begin{equation}
\label{equ:solve_dp}
\mathbf{d}_p=\mathbf{R}_{pp}^{-1}(\lambda_{h}\mathbf{Z}_p-\mathbf{R}_{pf}\mathbf{d}_f).
\end{equation}

As is noted above, we temporarily ignore all the inequality constraints and derive the formulation as an unconstrained QP.
Given time durations of trajectory segments and the weights of cost terms, the solution that minimizes the overall cost can be obtained efficiently in closed-form.
To ensure the feasibility of the final solution, after solving Eq.~\ref{equ:solve_dp}, we check whether safety and dynamical feasibility constraints are violated in each iteration, as shown in Alg.~\ref{alg:optimize_process}.
This is done by an extremely efficient continuous-time feasibility checker proposed in~\cite{wang2020alternating}. With our kinodynamic front-end providing initial trajectories with proper time allocation, the feasibility constraints are prone to be satisfied, as shown in our experimental tests.

\subsection{Optimization Process}

The optimization process is shown in Alg. \ref{alg:optimize_process}.
\begin{algorithm}[t]
\caption{Iterative Optimization}
\label{alg:optimize_process}
\begin{algorithmic}[1]
\State \textbf{Notation}: Environment $\mathcal{E}$, Trajectory $\pi$
\State Initialize: $r_c \leftarrow r_{c,init}$, $r_h \leftarrow r_{h,init}$, $\pi \leftarrow \pi_{init}$
\While{$r_c > 0$}
    \State $\pi_{temp} \leftarrow$ \textbf{ClosedFormSolve}($r_c$, $r_h$)
    \State $\pi \leftarrow \pi_{temp}$
    \If{$\neg$ \textbf{CheckFeasible}($\pi$, $\mathcal{E}$)} break;
    \EndIf
    \State $r_c \leftarrow r_c-d_{r,c}$
\EndWhile
\While{$r_h > 0$}
    \State $\pi_{temp} \leftarrow$ \textbf{ClosedFormSolve}($r_c$, $r_h$)
    \State $\pi \leftarrow \pi_{temp}$
    \If{$\neg$ \textbf{CheckFeasible}($\pi$, $\mathcal{E}$)} break;
    \EndIf
    \State $r_h \leftarrow r_h-d_{r,h}$
\EndWhile
\State \Return $\pi$
\end{algorithmic}
\end{algorithm}
Denote $r_c=\lambda_c/(\lambda_s+\lambda_h+\lambda_c)$ and $r_h=\lambda_h/(\lambda_s+\lambda_h)$.
The initial value of $r_c$ is set close to $1$ to prefer continuous acceleration between segments, and the initial value of $r_h$ is also set close to $1$ to make the solution similar to the original feasible one.
As shown above, in each iteration of the first loop, $r_c$ is decreased by $d_{r,c}$ while $r_h$ is fixed, and a temporary trajectory is obtained by \textbf{ClosedFormSolve()} with Eq.\ref{equ:solve_dp}, once this temporary trajectory is checked infeasible, the iteration stops and $r_c$ is fixed.
In this way, the segment-wise acceleration discontinuity is heavily penalized, and we will obtain a solution with near $C^2$ continuity.
In the second loop, $r_c$ is fixed and $r_h$ is decreased by $d_{r,h}$ in each iteration.
As $r_h$ continues to decrease, the importance of the smoothness term increase. Thus it seeks a smoother trajectory in a relatively small solution space around the solution of the same homotopy class provided by the original trajectory, meanwhile satisfying the feasibility checking.

\section{Benchmark Comparisons}
\subsection{Sampling Strategy}
\begin{figure}[t]
\centering
\includegraphics[width=0.95\linewidth]{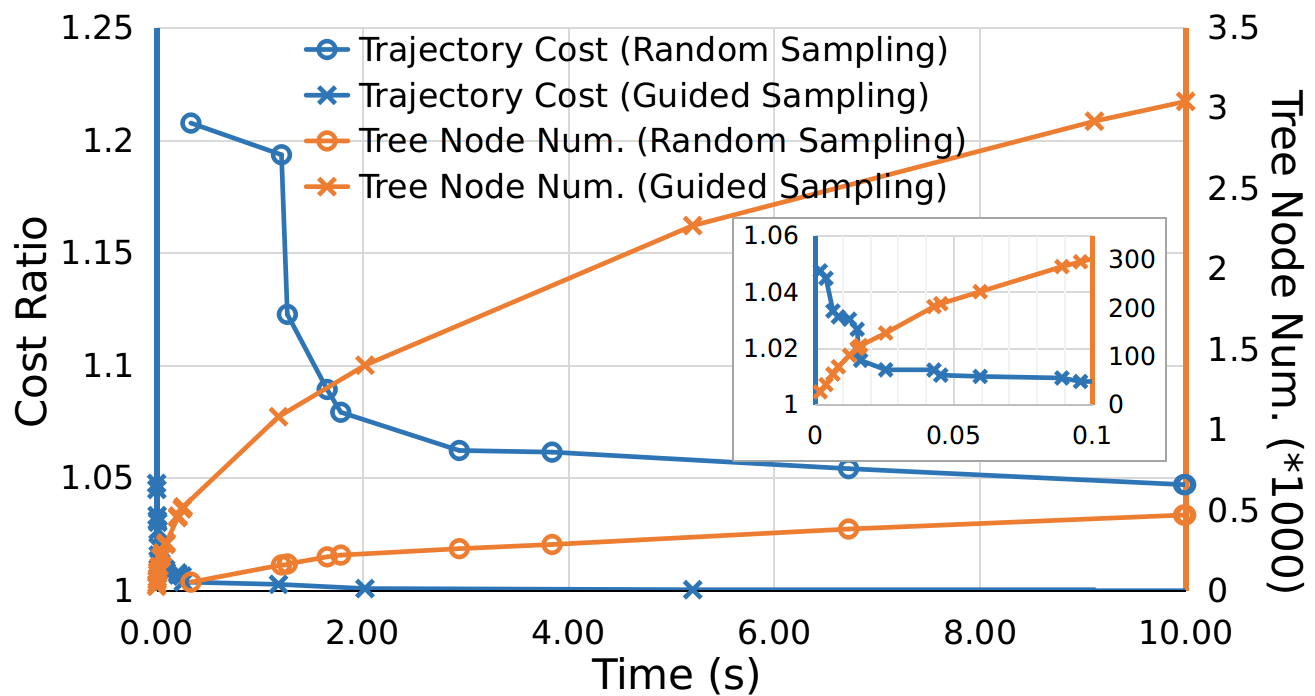}
\captionsetup{font={small}}
\caption{Comparison of our guided sampling strategy and random sampling method. The inner figure shows the detailed result between 0 to 0.1s.}
\label{pic:guided_sampling_comparison}
\vspace{-0.7cm}
\end{figure}

We compare our proposed topology guided sampling strategy with typical uniform sampling.
We conduct a random simulation in a $40\times40\times3m$ environment with 100 randomly deployed obstacles and starting and goal positions.
All the benchmark computations are done with a 2.2GHz Intel i7-4702HQ processor.
We limit the maximum planning time to $10s$ and take the trajectory cost of our method as a baseline.
The optimality ratio against planning time is shown in Fig.~\ref{pic:guided_sampling_comparison}.

As shown in Fig.~\ref{pic:guided_sampling_comparison}, using our guided sampling strategy, the cost decreases rapidly after the first solution found within a few milliseconds.
In contrast, it takes hundreds of milliseconds to find the first trajectory with uniformly random sampling, and the cost takes much longer time to approach the optimum.
As also validated in this figure, given a time budget, our method generates more feasible samples due to the reasonable state distribution.
Therefore, our method has a higher possibility of accessing a better solution and converges faster.
Fig.~\ref{pic:guided_vs_random} presents an illustrative sample of the comparison.


\subsection{Quadrotor Planning System}
We conduct benchmark comparisons against the state-of-the-art quadrotor online planning methods in three-folds: the front-end kinodynamic planning, the back-end trajectory optimization, and the integrated systematic results.
Simulations are conducted in environments with different obstacle densities and starting-goal distances.
The velocity and acceleration limits are set as $5m/s$ and $6m/s^2$.

\subsubsection{Comparisons of the Kinodynamic Planning}

For the front-end, the first feasible trajectory found by our method is compared against Zhou's.
As shown in Tab.~\ref{tab:front-end}, our method finds trajectories with much lower control costs, shorter trajectory length, higher success rate, and comparable computing time.
Since our method generates properly distributed state samples and explores the environment according to the topological structure, it finds a solution with fewer states expanded.
Besides, it better exploits the results of BVPs and generates piece-wise linear inputs instead of the piece-wise constant ones of method~\cite{boyu2019ral}, thus improves the smoothness.
Although our method has slightly higher computing time, it provides a much better initial trajectory and thus significantly alleviates the computational burden of the back-end, as validated below.

\subsubsection{Comparisons of the Trajectory Optimization}
\begin{figure}[t]
\centering
	\includegraphics[width=0.9\linewidth]{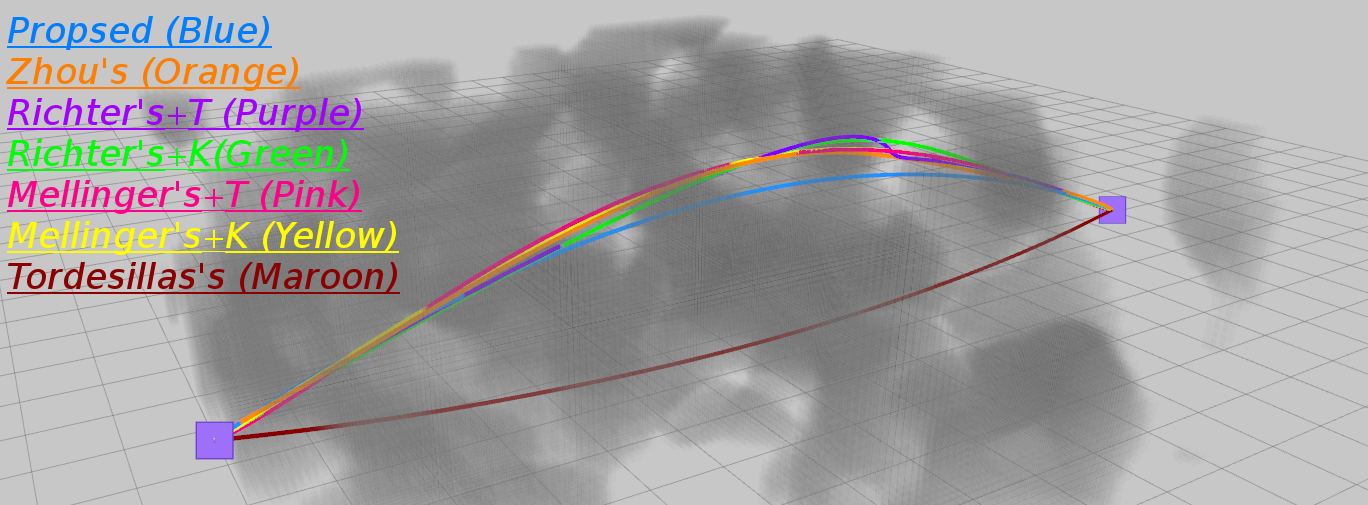}
\captionsetup{font={small}}
\caption{An instance of trajectories generated by different methods. The 3D obstacles are set transparent to provide better views.}
\label{pic:back_end_compare_add}
\end{figure}

For fair comparisons, we use the same path returned by our front-end planner as the initial value for Zhou's \cite{boyu2019ral} that adopts a B-spline formulation and optimize control points in a distance field with gradient descent to ensure safety, Richter's \cite{RicBryRoy1312} that optimizes derivatives on waypoints through an unconstrained QP while adjusting time allocation by gradient descent and scaling, and Mellinger's \cite{MelKum1105} that optimizes time allocation with total duration fixed and use backtracking gradient descent. 
For Tordesillas's\cite{tordesillas2019faster}, front-end paths are found by informed-RRT*.  
For waypoints-based methods Richter's and Mellinger's, both trapezoidal time initialization and the time allocation produced by our kinodynamic front-end are used and compared (denoted as T and K, respectively). 
When collisions occur in a particular segment, a point from the collision-free front-end path is added as an additional waypoint. The trajectory is then re-optimized, and the process is repeated until the whole trajectory is collision-free.
The stopping criterion for each iteration is set as $5ms$ running time.
The results are shown in Tab.~\ref{tab:back-end} and Fig.~\ref{pic:back_end_compare_add}.
Our proposed method generates much smoother and shorter trajectories in much less time.
This is because the compared ones adopt an underlying non-convex gradient-based formulation and require expensive computations for a general nonlinear optimization solver to converge.
Our method, however, enjoys the convex formulation to find the optimal solutions in its every iteration. 
\cite{boyu2019ral} checks collision in an ESDF which requires extra computation ($50ms$ in the testing case). \cite{tordesillas2019faster} builds a free corridor and perform MIQP with it. \cite{RicBryRoy1312} and \cite{MelKum1105} do not account for collision in optimization thus may require many iterations to find a collision-free trajectory. 
Our method, however, avoids these by incorporating the collision-free front-end trajectory into the objective.
Note that although the proposed method does not optimize time allocation, our final trajectory duration is less than Richter's and Mellinger's that use trapezoidal initialization, and is comparable to the ones that use the time allocation of our kinodynamic front-end.

\begin{table}[t]
\centering
\captionsetup{font={small}}
\caption{Front-end Comparison Results of 10-15m Goals in 150 Obstacles Environment.}
\label{tab:front-end}
\begin{tabular}{|c|c|c|c|c|c|c|}
\hline
Method & \begin{tabular}[c]{@{}c@{}}Comp.\\ Time\\ (ms)\end{tabular} & \begin{tabular}[c]{@{}c@{}}Seg.\\ Num.\end{tabular} & \begin{tabular}[c]{@{}c@{}}Ctrl.\\ Cost\\ ($m^2/s^3$)\end{tabular} & \begin{tabular}[c]{@{}c@{}}Traj.\\ Dura.\\ (s)\end{tabular} & \begin{tabular}[c]{@{}c@{}}Traj.\\ Len.\\ (m)\end{tabular} & \begin{tabular}[c]{@{}c@{}}Succ.\\ Rate\\ ($\%$)\end{tabular} \\ \hline
\begin{tabular}[c]{@{}c@{}}Proposed\\ First Traj.\end{tabular} & 4.76 & \textbf{4.05} & \textbf{24.97} & 5.49 & \textbf{13.21} & \textbf{96.01} \\ \hline
Zhou's & 4.58 & 6.50 & 40.16 & 5.33 & 13.45 & 94.07 \\ \hline
\end{tabular}
\end{table}

\begin{table}[t]
\centering
\captionsetup{font={small}}
\caption{Back-end Comparison Results.}
\label{tab:back-end}
\begin{tabular}{|c|c|c|c|c|c|}
\hline
Method & \begin{tabular}[c]{@{}c@{}}Comp.\\ Time\\ (ms)\end{tabular} & \begin{tabular}[c]{@{}c@{}}Inte. of\\ Acc.\\ ($m^2/s^3$)\end{tabular} & \begin{tabular}[c]{@{}c@{}}Inte. of\\ Jerk\\ ($m^2/s^5$)\end{tabular} & \begin{tabular}[c]{@{}c@{}}Traj.\\ Dura.\\ (s)\end{tabular} & \begin{tabular}[c]{@{}c@{}}Traj.\\ Len.\\ (m)\end{tabular} \\ \hline
Proposed & \textbf{2.82} & \textbf{18.72} & \textbf{36.17} & {5.42} & 12.96 \\ \hline
{Zhou's} & 6.20 & 19.16 & 92.80 & 5.42 & 13.17 \\ \hline
{Richter's + T} & {24.79} & {77.94} & {505.95} & {5.74} & {13.22} \\ \hline
{Richter's + K} & {17.39} & {34.17} & {197.06} & {5.42} & {13.18} \\ \hline
{Mellinger's + T} & {26.92} & {34.53} & {78.77} & {6.01} & {13.19} \\ \hline 
{Mellinger's + K} & {32.12} & {29.60} & {98.60} & {5,13} & {13.18} \\ \hline
{Tordesillas's} & {183.03} & {24.52} & {45.51} & \textbf{4,97} & {\textbf{12.57}} \\ \hline
\end{tabular}
\end{table}

\subsubsection{Comparisons of the Integrated Results}
\begin{figure}[htbp]
\centering
    \begin{subfigure}[hb]{0.9\linewidth}
        \includegraphics[width=\linewidth]{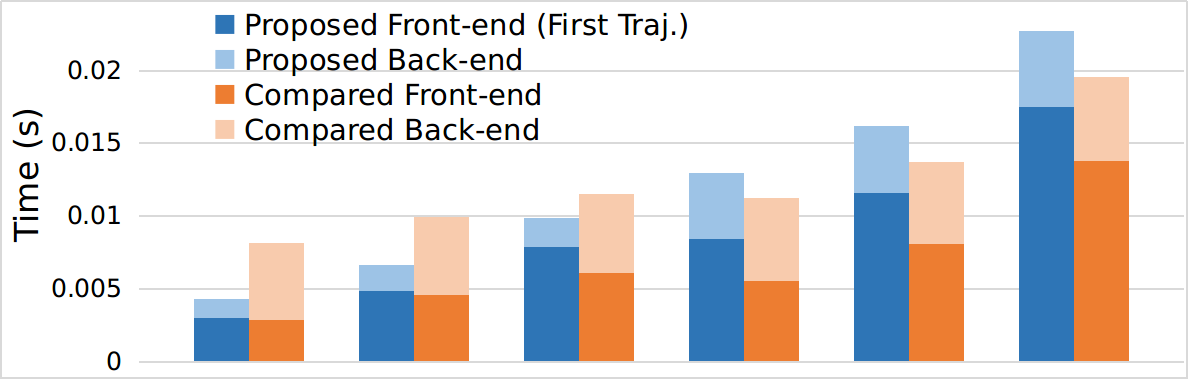}
        \caption{Planning time ($ms$)}
        \label{subfig:time}
    \end{subfigure}
    \begin{subfigure}[hb]{0.9\linewidth}
        \includegraphics[width=\linewidth]{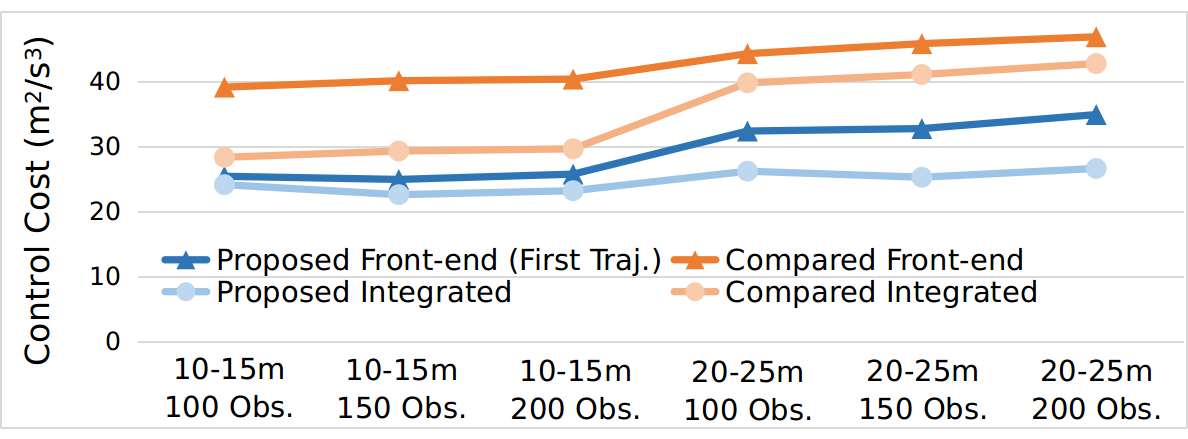}
        \caption{Control effort cost ($m^2/s^3$)}
        \label{subfig:control_cost}
    \end{subfigure}
\captionsetup{font={small}}
\caption{Comparison of the integrated results in environment of different obstacle densities and different distance goals.}
\label{fig:front_end_integration}
\end{figure}

For the integrated comparison, results of different scenarios are shown in Fig.~\ref{fig:front_end_integration}.
As an entire planning pipeline, our system generates trajectories with much lower control cost in each scenario and less time used in relatively short distances.
However, as the goal distance and obstacle density increase, our method requires a bit more time than Zhou's method.
For a planning problem with a large scale, samples near the goal are inferior to grow the tree since they are less likely to safely connect to an existing state, especially in complex environments.
However, this is not critical since for common real-world applications, the sensing range and planning horizon of a lightweight drone are usually within $10m$, or even $5m$.
It is verified in our real-world tests in Sec.~\ref{sec:results}.

\section{Flight Experiments}
\label{sec:results}
\subsection{Experiment Settings}
We conduct autonomous flight experiments in both indoor and outdoor unknown cluttered environments.
The flight platform we use is a customized quadrotor equipped with a forward-facing RealSense D435i\footnote{https://www.intelrealsense.com/depth-camera-d435i/} and an N3 flight controller\footnote{https://www.dji.com/cn/n3} for depth sensing and flight control.
Collision checking is done with occupancy grid maps~\cite{thrun2005probabilistic} fused by the depths and the poses estimated.
Unknown space is treated as free. Replan is conducted when obstacles are newly perceived or new goals are set.
All the computations are done online with an onboard computer Manifold2-C\footnote{https://www.dji.com/cn/manifold-2}. 

\subsection{Waypoints Navigation}
\begin{figure*}[h]
\centering
\begin{subfigure}{0.25\linewidth}
	\includegraphics[width=1\linewidth]{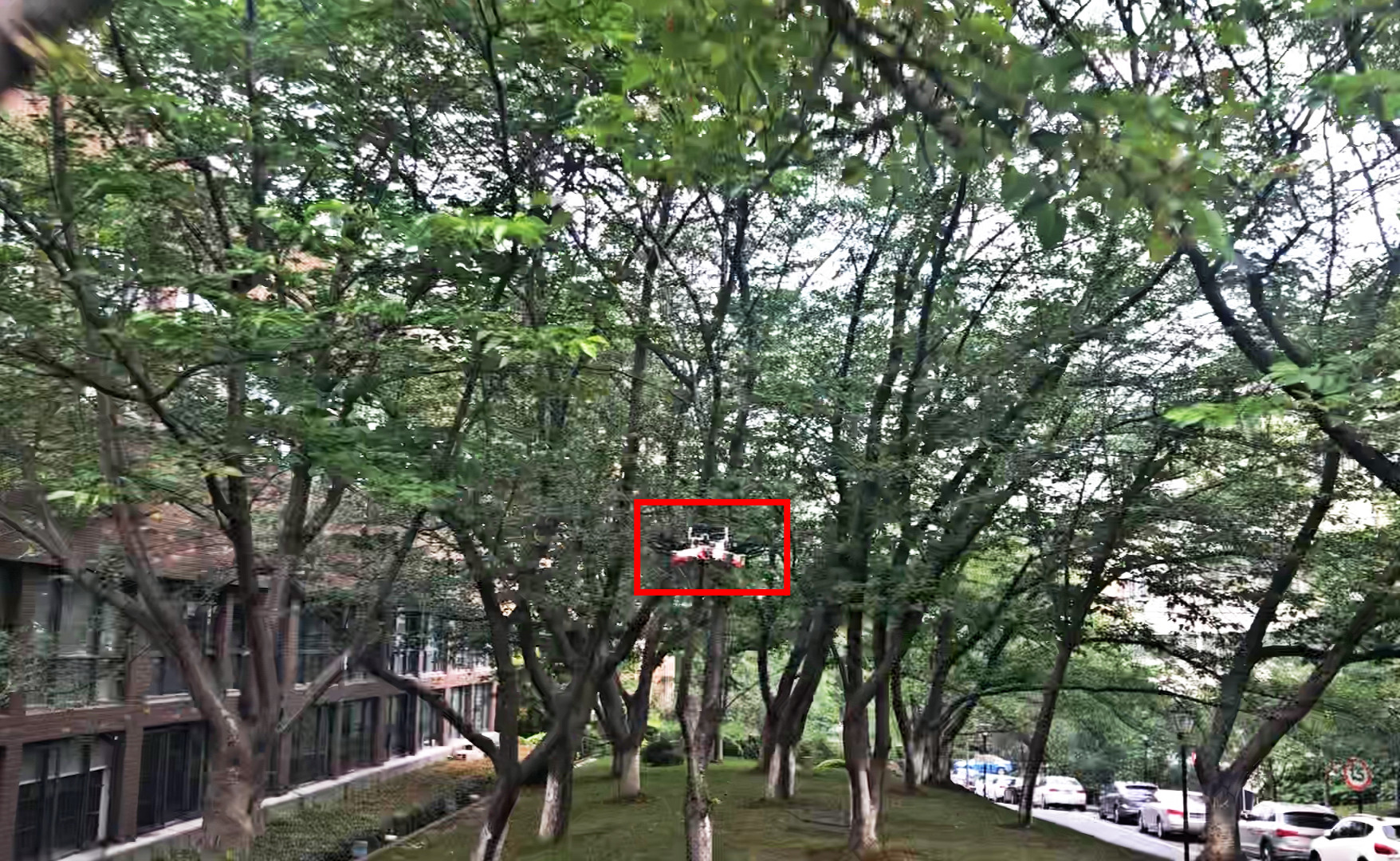}
	\caption{}
\end{subfigure}
\begin{subfigure}{0.74\linewidth}
	\includegraphics[width=1\linewidth]{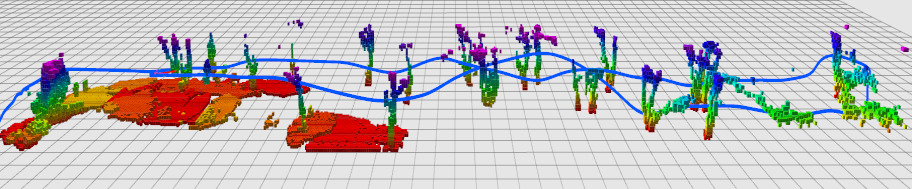}
	\caption{}
\end{subfigure}
\vspace{-0.15cm}
\captionsetup{font={small}}
\caption{Outdoor flights. The quadrotor flies about $100m$ with an average speed of about $3m/s$. Velocity profiles are plotted in the video.}
\label{pic:outdoor_flight}
\vspace{-0.2cm}
\end{figure*}

\begin{figure}[h]
\centering
\begin{subfigure}{0.415\linewidth}
	\includegraphics[width=1\linewidth]{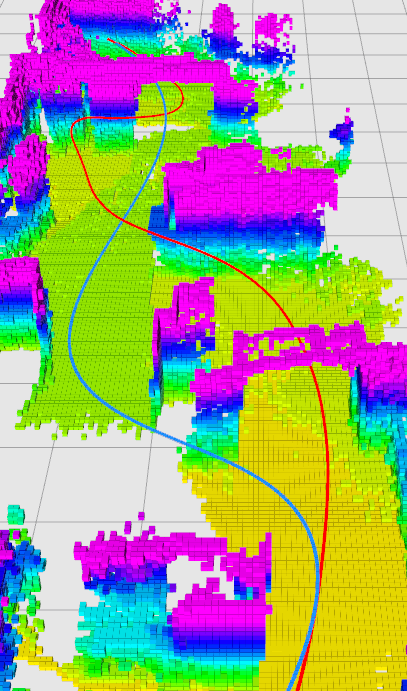}
	\caption{}
\end{subfigure}
\begin{subfigure}{0.55\linewidth}
	\includegraphics[width=1\linewidth]{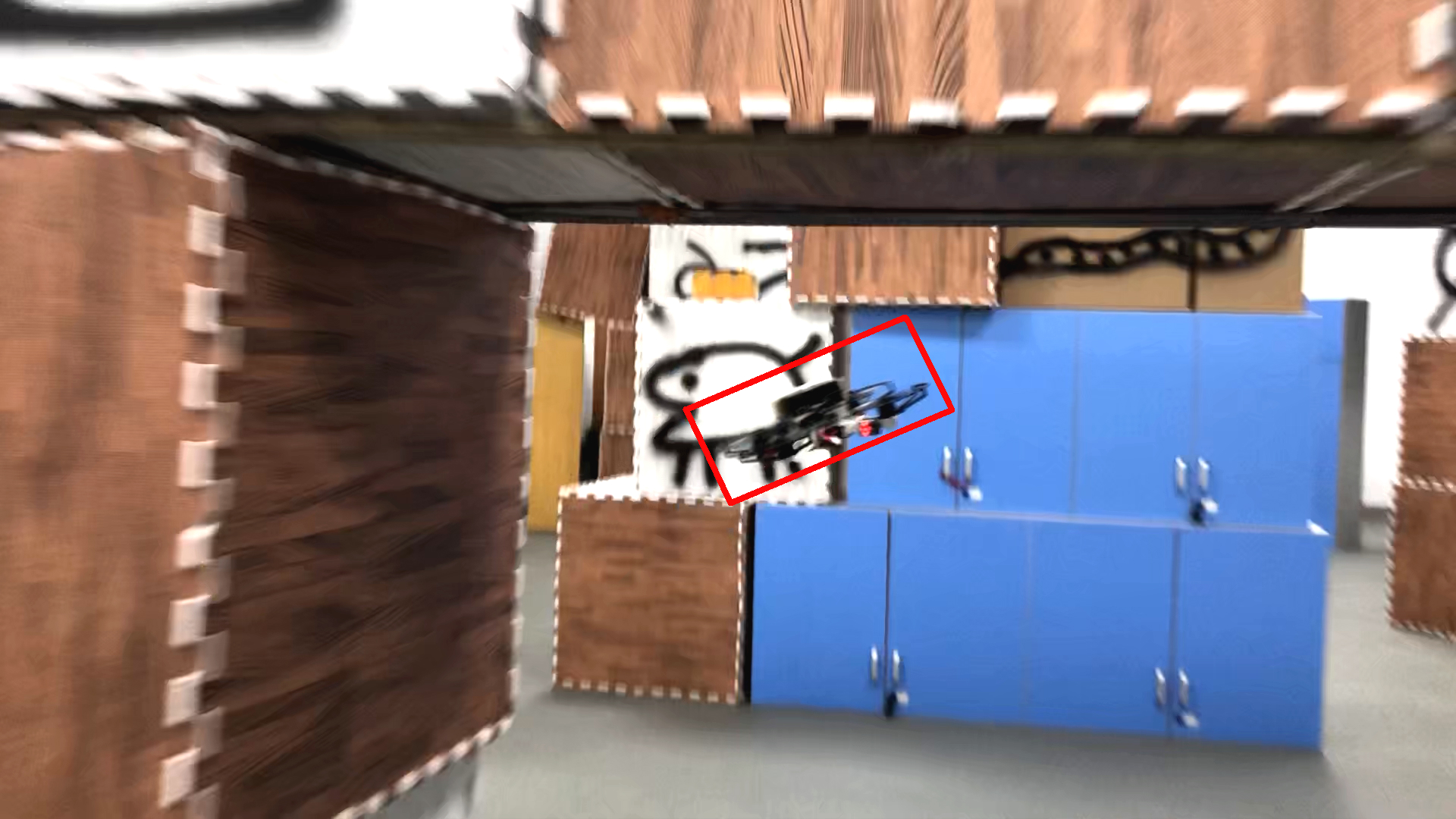}
	\caption{}
	\includegraphics[width=1\linewidth]{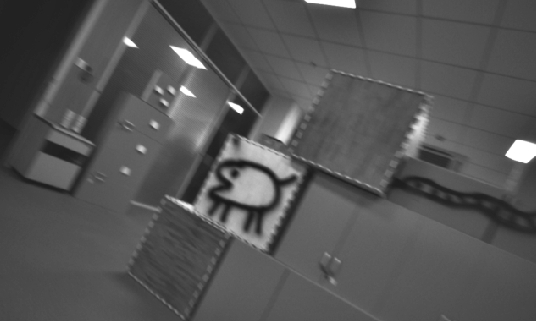}
	\caption{}
\end{subfigure}
\vspace{-0.15cm}
\captionsetup{font={small}}
\caption{Indoor flights. (a) The trajectory planned back (blue) is smoother than the departure trajectory (red) since it has seen part of the environment in the previous flight. (b) The quadrotor makes a turn when facing a wall right after flying through a gate. The average speed is about $2.5m/s$. (c) The first person view.}
\label{pic:indoor_flight}
\vspace{-0.4cm}
\end{figure}

The quadrotor, with limited sensing range ($3m$) and field of view ($60^{\circ}$), navigates to a goal of about $50m$ and $15m$ away and then come back in the outdoor and indoor flight tests, respectively.
The executed trajectories are depicted in Fig.~\ref{pic:outdoor_flight} and Fig.~\ref{pic:indoor_flight}.
In the outdoor flight, the quadrotor operates in previously unknown dense and unstructured woods.
In the indoor environment, the obstacles are more massive and cause more occlusions. Thus some obstacles are more likely to appear suddenly.
In these experiments, our planner shows its capability to facilitate autonomous navigation while avoiding obstacles.
More details are available in the video.

\subsection{Fast Replan Tasks}
To further challenge our planner and test the replan performance, we conduct tasks with continually changing goals for the quadrotor to chase in unknown cluttered woods. Replan happens whenever the goal changes or the current tracking path is blocked by a newly detected obstacle.
In the first task, the quadrotor is made to chase after a fast-moving target, a QR code board, which determines the goal position (See Fig.~\ref{fig:chase_qr}).
In the second task, the goals are set and changed arbitrarily and abruptly at any time during flight by an operator.
Our drone keeps a speed over $3m/s$ while planning new trajectories as soon as newly observed obstacles block the current flight trajectory.
Higher speed can be achieved with longer confidence sensing range and less latency, which is mainly caused by map fusion.
We refer readers to the video for more flight tests.

\section{Conclusion}
\label{sec:conclusion}
In this paper, a novel online motion planning framework for quadrotor fast flight is proposed.
The method is composed of 1) a guided sampling-based kinodynamic planner for finding an initial safe, kinodynamiclly feasible and time-energy optimal trajectory and 2) a homotopy penalized, soft constrained, iterative optimizer to further improve the smoothness and continuity of the trajectory.
Benchmark comparisons show that our method outperforms the state-of-the-art methods in both efficiency and optimality.
Moreover, we validate our method in simulated and real-world challenging tasks.
In the future, we plan to further improve the obstacle clearance of the refined trajectory and challenge our method for large-scale problems.

\newlength{\bibitemsep}\setlength{\bibitemsep}{.00\baselineskip}
\newlength{\bibparskip}\setlength{\bibparskip}{0pt}
\let\oldthebibliography\thebibliography
\renewcommand\thebibliography[1]{%
  \oldthebibliography{#1}%
  \setlength{\parskip}{\bibitemsep}%
  \setlength{\itemsep}{\bibparskip}%
}
\bibliography{main}
\end{document}